# Streamline pathology foundation model by cross-magnification distillation

Ziyu Su[a*], Abdul Rehman Akbar[a†*], Usama Sajjad[a†], Anil V. Parwani[a], Muhammad Khalid Khan Niazi[a]

[a]Department of Pathology, College of Medicine, The Ohio State University Wexner Medical Center, Columbus, OH,USA

*Correspondence: ziyus21@outlook.com (Z.S.), Abdul.Akbar@osumc.edu (A.R.A.)

†These authors contributed equally to this work.

## Abstract

Foundation models (FM) have transformed computational pathology but remain computationally prohibitive for clinical deployment due to their massive parameter counts and high-magnification processing requirements. Here, we introduce XMAG, a lightweight FM developed through cross-magnification distillation that transfers knowledge from state-of-the-art 20x magnification teacher to an efficient 5x magnification student architecture. XMAG employs a compact backbone and operates entirely at 5x, requiring 11.3 times fewer patches per whole slide image (WSI) compared to existing approaches. Our novel distillation framework incorporates dual-level knowledge transfer, aligning both global image representations and local spatial features across magnification levels through carefully designed projection heads and spatial token mapping. We trained XMAG on 3.49 million images curated from publicly available datasets and evaluated performance across seven clinically relevant histopathology analysis tasks spanning multiple cancer types. XMAG achieved diagnostic accuracy within 1% of substantially larger foundation models while delivering 30-fold encoding acceleration, reaching 8.8 WSIs per minute encoding speed. We further explore end-to-end XMAG training as a direction for task-specific adaptation with compact model. These results suggest that cross-magnification distillation is a promising strategy for developing efficient low-magnification pathology foundation models and may support more scalable WSI analysis in computationally constrained research and translational settings.

## 1. Introduction

Computational pathology (CPath) has emerged as a cornerstone of modern diagnostic medicine[1, 2], yet a fundamental challenge persists: achieving both high diagnostic accuracy and the rapid processing speeds required for clinical integration. Current CPath workflows [3-8] face a critical bottleneck where the time required for AI analysis often exceeds the pace of clinical decision-making, particularly in time-sensitive applications such as intraoperative consultations and high-throughput screening programs where pathologists require near real-time feedback[9].

Foundation models (FMs) have revolutionized CPath by achieving unprecedented diagnostic accuracy across diverse histopathological tasks [3-5]. However, state-of-the-art models operate at high magnification (40-20x) with massive parameter counts—UNI2 [3] (682M parameters), Virchow2 [10] (632M parameters), and Gigapath [5] (1.1B parameters)—creating substantial computational demands that effectively preclude clinical deployment. These computational requirements are particularly problematic for on-premises deployment in hospital environments, where data privacy regulations and institutional policies often mandate local processing of sensitive patient information on resource-constrained infrastructure.

The computational challenge is fundamentally rooted in a magnification-efficiency trade-off. Higher magnification imaging captures more detailed morphological features but requires exponentially more patches for analysis—20x magnification necessitates processing 16-fold more patches than 5x magnification for the same tissue area (see Figure 1a) [11]. Current approaches assume that high magnification is essential for competitive diagnostic performance, raising a critical question: can comparable accuracy be achieved at lower magnification using more efficient architectures? [12, 13]

Knowledge distillation presents a promising avenue for addressing these challenges by enabling compact models to learn from larger, more capable teachers [14-16]. However, no framework currently exists for cross-magnification knowledge transfer in CPath—an approach that could potentially combine the detailed feature representations learned at high magnification with the computational efficiency of low-magnification analysis. Such a framework could not only improve processing efficiency but also overcome data limitations by transferring learned representations from models trained on larger datasets [17].

Here, we introduce XMAG (Cross-Magnification Distillation), a novel framework that transfers knowledge from a 20x magnification teacher model to a compact 5x magnification student model with only 86M parameters. XMAG operates entirely at 5x magnification during inference. Theoretically, reducing the magnification from 20x to 5x decreases the number of patches required to cover the same tissue area by 16-fold, because each 5x patch covers a 4x larger field of view along each spatial dimension. In our benchmark datasets, this translated into an empirical average reduction of 11.3-fold in patch counts compared with 20x processing. The compact architecture and reduced patch requirements enable end-to-end training for WSI-level classification [18], allowing task-specific adaptation that is previously challenging with larger models due to memory constraints.

Remarkably, we systematically assembled and curated a training dataset of 3.49 million patches span across sixteen different cancer types from publicly available sources. XMAG achieves performance comparable to state-of-the-art FMs trained on larger private datasets [3], demonstrating that cross-magnification distillation enables efficient knowledge transfer that compensates for dataset scale differences. This combination of our curated public dataset and distillation framework democratizes access to high-performance pathology AI.

We demonstrate XMAG's capabilities across seven clinically relevant tasks, including four WSI-level classification tasks, two patch-level phenotyping tasks, and one lesion segmentation task. These tasks span multiple cancer types and multiple hospitals, supporting the generalizability of XMAG. XMAG achieved an encoding speed of 8.8 WSIs per minute, representing an 8–30-fold improvement over existing foundation models in the controlled patch-processing benchmark, while maintaining performance close to substantially larger 20x models across several tasks. These results suggest that cross-magnification distillation is a promising strategy for developing efficient low-magnification pathology foundation models and may support scalable WSI analysis in computationally constrained research and translational settings.

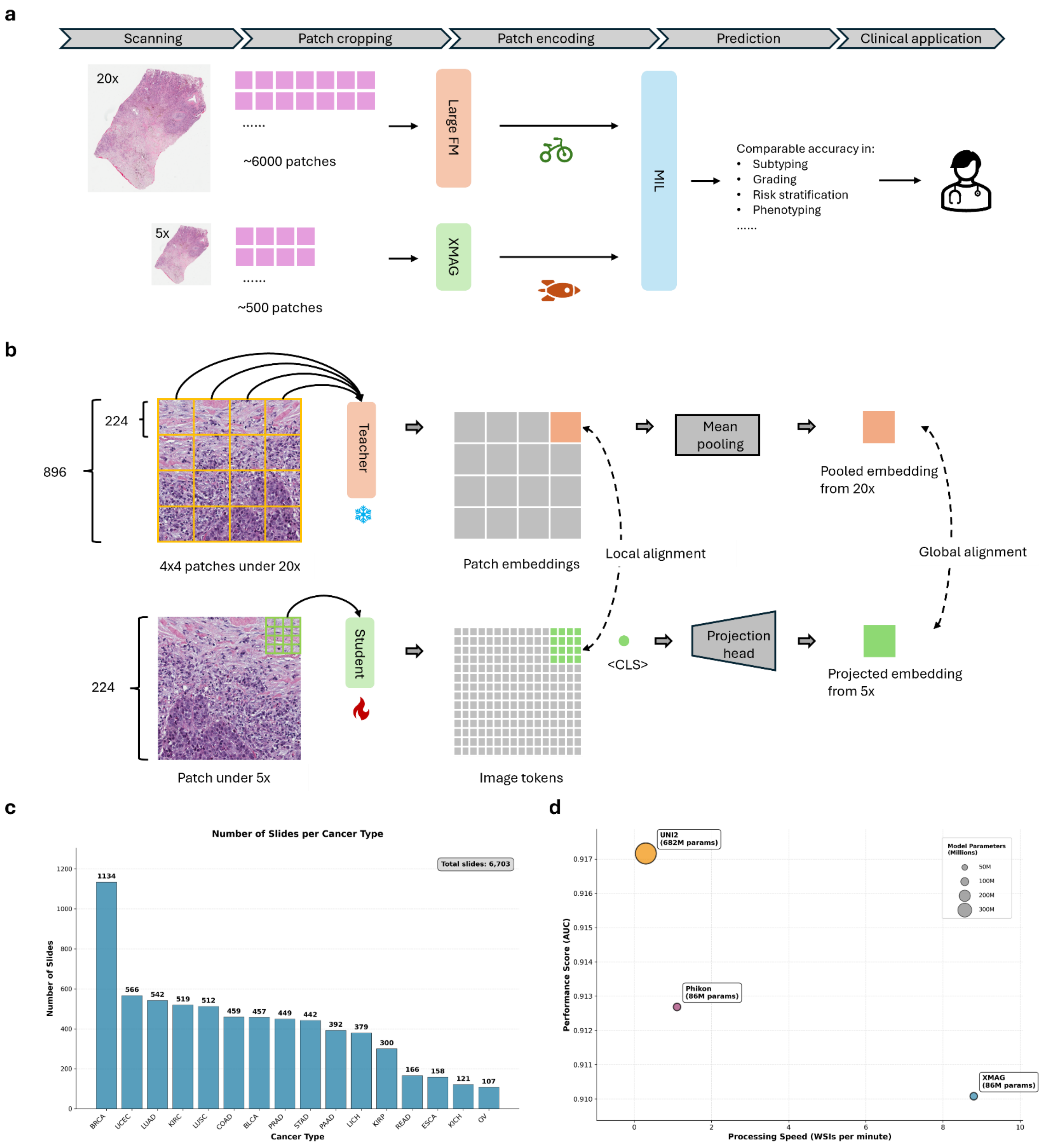

Figure 1. Overview of the XMAG. XMAG transfers knowledge from high-magnification foundation models to compact low-magnification architectures, achieving comparable diagnostic performance with dramatically improved processing efficiency across multiple clinical tasks. **a.** Empirical patch-count reduction across four WSI benchmark datasets. Traditional 20x approaches required an average of 6,260 patches per WSI, whereas XMAG operated at 5x magnification with an average of 554 patches per WSI, corresponding to an 11.3-fold reduction in patch count. **b.** Cross-magnification distillation architecture. Teacher model processes 896×896 patches as 4×4 grids of sub-patches, generating embeddings for global and local alignment. Student model processes 224×224 patches with spatial token mapping and projection heads for dual-level knowledge transfer. **c.** Our pretraining dataset comprises 6,703 WSIs across 16 cancer types from public sources,

totaling 3.49 million patches for distillation training. **d.** XMAG achieves optimal balance between diagnostic performance (average AUC) and processing speed (8.8 WSIs/minute) compared to existing foundation models. Bubble size indicates parameter count.

## 2. Methods

### 2.1. Pretraining dataset

We established a cross-magnification pretraining dataset with about 3.49 million patches under 5x magnification from hematoxylin and eosin (H&E) WSIs. The dataset is a multi-cancer dataset created from the publicly available WSIs from TCGA[1] (see Figure 1c). It incorporates sixteen different cancer types and 6703 WSIs. Based on our distillation approach, we require each 5x patch has its equivalent 20x region available. Thus, when creating the dataset, we cropped patches in 896×896 pixel size under 20x magnification and then downscale them into 224×224 to create the 5x patches. Before resizing, we divide the 896×896 patch into 4 by 4 grids to generate sixteen 224×224 patches under 20x. In summary, each 5x patch corresponds to sixteen 20x patches during the cross-magnification distillation.

### 2.2. Self-supervised training using cross-magnification distillation

Knowledge distillation is a technique in which a lightweight student model learns to align its feature representations with those of a more powerful teacher model [15]. Consequently, the student can achieve similar performance as the teacher model does with fewer parameter. To enable knowledge transfer from high-magnification foundation models to computationally efficient low-magnification architectures, we developed a novel cross-magnification distillation framework that addresses the fundamental challenge of aligning features across different spatial resolutions and magnification levels. Specifically, our method establishes correspondence between 16 non-overlapping 20× patches and a single spatially-equivalent 5× patch, enabling the distillation of fine-grained morphological knowledge into a more efficient representation. Our model architecture is outlined in Figure 1b.

#### *2.2.1. Teacher-Student Architecture*

Let $X^{20\times} \in \mathbb{R}^{896\times896\times3}$ denote a high-magnification patch at 20× and $X^{5\times} \in \mathbb{R}^{224\times224\times3}$ represent the corresponding low-magnification patch at 5x, where both patches capture the same tissue region. Our teacher network is UNI2, a state-of-the-art FM pretrained with 100 million patches of 256×256 under 20x. It applies a ViT-h backbone with 682 million parameters. The teacher network $f_T$ processes $X^{20\times}$ by first decomposing it into 16 non-overlapping sub-patches:

$$X^{20\times} = \{x_i^{20\times}\}_{i=1}^{16}, \qquad x_i^{20\times} \in \mathbb{R}^{224\times224\times3}$$

Each sub-patch is independently encoded to produce high-dimensional patch features:

$$h_i^T = f_T(x_i^{20\times}) \in \mathbb{R}^{d_T}, \qquad i = 1, \ldots, 16$$

where $d_T = 1536$ is the teacher embedding dimension.

[1] https://portal.gdc.cancer.gov/

The student network is based on the DINOv2-ViT-B[19] architecture and contains 86 million parameters. The student network $f_s$ processes the corresponding 5× patch through a single forward pass, yielding a class token and spatial patch tokens:

$$c^S, \{t_j^S\}_{j=1}^{256} = f_S(X^{5\times})$$

where $c^S \in \mathbb{R}^{d_S}$ is the global class token and $t_j^S \in \mathbb{R}^{d_S}$ are patch tokens with $d_S = 768$.

*2.2.2. Spatial Resolution Alignment*

To establish spatial correspondence between teacher and student representations, we perform spatial pooling on the student's 256 patch tokens, which are arranged in a 16×16 spatial grid. We reshape and aggregate these tokens into 16 spatially-aligned features through 4×4 average pooling:

$$T^S = reshape\left(\{t_j^S\}_{j=1}^{256}, 16 \times 16\right) \in \mathbb{R}^{16\times16\times d_S}$$

$$h_i^S = \frac{1}{16}\sum_{p=1}^{4}\sum_{q=1}^{4} T_{4i_r+p,4ic+q}^S\ , \qquad i = 1, \dots, 16$$

where $i_r$ and $i_c$ denote the row and column indices corresponding to the $i$-th spatial region (see Figure 1b).

*2.2.3. Dimension Matching through Projection Heads*

To align the dimensional disparity between teacher ($d_T = 1536$) and student ($d_S = 768$) representations, we employ learnable projection heads. For global features, a two-layer MLP $g_{global}$ projects the student class token to 1536 dimensional space:

$$z_{global}^S = g_{global}(c^S) = W_2 \ \cdot GELU\big(BN(W_1 c^S + b_1)\big) + b_2$$

where $W_1 \in \mathbb{R}^{d_T\times d_S}$, $W_2 \in \mathbb{R}^{d_T\times d_T}$, and $BN$ denotes batch normalization.

Similarly, for local features, each spatially-aligned student feature is projected to 1536 dimensional space:

$$z_i^S = g_{local}\big(h_i^S\big), \qquad i = 1, \dots, 16$$

*2.2.4. Knowledge Distillation*

Our distillation objective combines global and local alignment losses. The global teacher representation is obtained from by averaging spatial features:

$$h_{global}^T = \frac{1}{16}\sum_{i=1}^{16} h_i^T$$

The global distillation loss employs negative cosine similarity:

$$L_{global} = -\frac{h_{global}^T \cdot z_{global}^S}{\| h_{global}^T \|_2 \, \| z_{global}^S \|_2}$$

The local distillation loss aligns spatially-corresponding features:

$$L_{local} = -\frac{1}{16}\sum_{i=1}^{16} \frac{h_i^T \cdot z_i^S}{\left\| h_i^T \right\|_2 \left\| z_i^S \right\|_2}$$

The total distillation loss is:

$$L = \lambda_{global} L_{global} + \lambda_{local} L_{local}$$

where $\lambda_{global} = 1$ and $\lambda_{local} = 0.5$ are weighting coefficients.

*2.2.5. Training details*

We employ data augmentation ensuring identical transformations are applied to both teacher and student inputs to maintain spatial correspondence. The student parameters are optimized using AdamW with learning rate $5 \times 10^{-4}$ and cosine annealing schedule. An exponential moving average (EMA) of student parameters is maintained to stabilize the training [20]. After training, the EMA student network is saved as our finalized XMAG model. Finally, our model was trained for 30260 iterations using 24 NVIDIA A100-40GB GPUs.

### 2.3. WSI-level classification

*2.3.1. Multiple Instance Learning*

For all WSI-level classification tasks, we employ MIL as the underlying framework. Each WSI is first tessellated into non-overlapping patches of 224×224 pixels at 5x magnification. We then utilize XMAG in inference mode to encode all patches from each WSI into patch-level embeddings:

$$h_i = f_{XMAG}\left(x_i^{5\times}\right), \qquad i = 1,2, \dots, M$$

where $f_{XMAG}(\cdot)$ denotes the XMAG, $h_i \in \mathbb{R}^{768}$ represents the patch embedding derived from XMAG's class token, $x_i$ is the $i$-th patch, and N indicates the total number of patches extracted from a WSI.

The resulting patch embeddings $\{h_i\}_{i=1}^{M}$ are then aggregated into WSI-level predictions using ABMIL [21], which learns to weight patch contributions based on their relevance to the classification task. During training, XMAG parameters remain frozen while only the ABMIL aggregation network is optimized for each downstream task.

*2.3.2. End-to-end training of MIL with XMAG*

Conventional MIL pipelines decouple FM inference from downstream task training due to GPU memory constraints. This separation limits the model's ability to learn task-specific features, as patch encodings are computed in a separate preprocessing step without gradient flow from the MIL objective. This architectural limitation prevents FMs from adapting their representations to specific downstream tasks during MIL training.

GPU memory consumption in end-to-end MIL training stems from two primary factors: (1) the large number of patches extracted from WSIs at high magnification, and (2) the substantial memory footprint of large FMs storing intermediate activations during backpropagation. XMAG presents a unique opportunity to address both limitations simultaneously. Operating at 5x magnification, XMAG naturally requires fewer patches compared to 20x models, reducing the spatial complexity of MIL training.

Additionally, XMAG's DINOv2-ViT-B backbone is substantially more compact than state-of-the-art pathology foundation models, resulting in significantly reduced memory requirements for intermediate activations.

Building on these advantages, we developed end-to-end XMAG (e2e-XMAG) that integrates XMAG with ABMIL in a unified training framework. To further optimize GPU memory usage, we employ gradient checkpointing to trade computation for memory by discarding intermediate activations during forward passes and recomputing them as needed during backpropagation.

Our implementation incorporates selective layer fine-tuning based on empirical ablation studies. We freeze the first ten ViT blocks of XMAG while allowing the final two blocks to adapt during MIL training. This selective unfreezing strategy, determined through systematic ablation experiments, provides an optimal balance between task-specific adaptation and computational efficiency. The frozen early layers maintain the robust low-level feature representations learned during cross-magnification distillation, while the trainable final layer enables task-specific feature refinement guided by the MIL objective.

The complete e2e-XMAG pipeline processes WSI patches through the partially-frozen XMAG encoder with gradient checkpointing, projects the resulting embeddings to the appropriate dimensionality for MIL aggregation, and jointly optimizes both the unfrozen XMAG parameters and ABMIL weights using the downstream classification loss. This end-to-end optimization enables the model to learn representations specifically tailored to each diagnostic task while maintaining the computational efficiency benefits of the cross-magnification distillation framework.

### 2.4. Statistical Methods

Statistical significance of model comparisons was assessed using paired testing procedures based on the task type and evaluation metric. For binary classification tasks, AUC was computed as the standard ROC-AUC and compared using DeLong's test for correlated ROC curves derived from paired predictions on the same test cases. Accuracy was compared using McNemar's test based on paired correctness indicators, where each case was coded as correctly or incorrectly classified by each model. F1-score was compared using paired bootstrap resampling at the slide/case level.

For multiclass classification tasks, AUC was computed as macro-averaged one-vs-rest AUC. Specifically, each class was treated as the positive class against all remaining classes, and the final AUC was obtained by averaging the class-wise AUCs. F1-score was reported as macro-F1, calculated as the mean of class-wise F1-scores. Because multiclass macro-AUC, overall accuracy, and macro-F1 are summary metrics that do not directly reduce to the paired binary outcomes required by McNemar's test or standard binary DeLong testing, statistical comparisons for these multiclass metrics were performed using paired bootstrap resampling at the slide level. For the segmentation experiments, Dice score was computed for each testing image. Model comparisons were performed using paired bootstrap resampling at the image level using the per-image Dice difference as the test statistic. Across all analyses, 95% confidence intervals were calculated using bootstrap resampling with 1,000 iterations. In Table S2, we summarized the statistical tests that we used for different tests.

## 3. Results

### 3.1. WSI-level classification

WSI classification represents the pinnacle of computational pathology applications, directly mirroring the diagnostic decision-making process that pathologists perform in clinical practice [22]. The ultra-high resolution of WSIs (typically exceeding $100{,}000 \times 100{,}000$ pixels) necessitates multiple instance learning (MIL) approaches, where WSIs are tessellated into manageable patches, encoded by FMs, and subsequently aggregated through MIL classifiers for slide-level predictions [6]. Due to GPU memory constraints, patch encoding and MIL classification operate as decoupled processes during training—FMs run in inference mode while only MIL aggregation layers undergo task-specific optimization. This architectural constraint places paramount importance on the representational quality of pretrained FMs, as their frozen feature extraction capabilities directly determine downstream classification performance without the ability to adapt to specific diagnostic tasks.

We benchmarked XMAG against two established pathology FMs using attention-based MIL (ABMIL) [21]: UNI2 [3], a state-of-the-art 682M parameter model, and Phikon [23], a more lightweight 86M parameter architecture. These two models were evaluated under their original 20x settings as practical reference benchmarks to assess how closely an efficient 5x model can approach the performance of established high-magnification foundation models while reducing computational cost. Also, we trained a 5x DINOv2 model [19] using our pretraining dataset and the same ViT-B backbone as XMAG. DINOv2 is also the core pretraining algorithm used by UNI/UNI2, although those models were trained on substantially larger pathology datasets, including private institutional data.

We selected the following tasks as our WSI-level benchmarks: non-small cell lung cancer (NSCLC) subtyping on the CPTAC-NSCLC dataset, renal cell carcinoma (RCC) subtyping on the DHMC-RCC dataset [24], prostate cancer ISUP grading on the PANDA challenge dataset [25], and breast cancer Oncotype-DX (BRCA-ODX) stratification on an internal OSUWMC-BRCA dataset. For all experiments, XMAG and DINOv2 used 224×224 patch size under 5x, and the other two models use 256×256 patch size under 20x magnification following their original implementations. For CPTAC-NSCLC and PANDA, we performed five-fold cross-validation and reported the average performance across folds. Since DHMC-RCC and OSUWMC-BRCA [26, 27] cohorts were relatively class-imbalanced, we treated them as independent testing cohorts and trained models using five-fold splits of TCGA-RCC and TCGA-BRCA [28], respectively. Each fold-specific model was then evaluated on the corresponding independent test cohort. Detailed descriptions of all datasets are provided in the *Datasets Used in the Study* section.

For NSCLC subtyping, we classified WSIs into lung adenocarcinoma (LUAD) and lung squamous cell carcinoma (LUSC). Under five-fold cross-validation, XMAG achieved an average area under the receiver operating characteristics curve (AUC) of $0.941 \pm 0.015$ (95% CI 0.922-0.942). For RCC subtyping, we performed three-way classification distinguishing papillary renal cell carcinoma (PRCC), chromophobe renal cell carcinoma (ChRCC), and clear cell renal cell carcinoma (ccRCC). XMAG achieved an average AUC of $0.986 \pm 0.009$ (95% CI 0.977-0.993) on the independent DHMC-RCC dataset, supporting cross-institutional generalization. For prostate cancer ISUP grading, we classified prostate biopsies into ISUP grades 0-5 [25]. Under five-fold cross-validation, XMAG achieved an average AUC of $0.910 \pm 0.003$ (95% CI 0.907-0.914). For BRCA-ODX stratification, we classified breast cancer cases into high or low recurrence risk based on Oncotype DX genetic test score thresholds (>25). XMAG achieved an average AUC of $0.776 \pm 0.006$ (95% CI 0.758-0.795) on the OSUWMC-BRCA dataset.

Remarkably, across most of the benchmark tasks, XMAG's performance was statistically indistinguishable (p>0.05) from the inefficient 20x FM (Figure 2). Only on the ISUP grading task did comparison models achieve significantly superior performance, with Phikon reaching an AUC of 0.923 (+0.013 over XMAG, p<0.001) and UNI2 achieving 0.925 (+0.015 over XMAG, p<0.001). Conversely, in RCC subtyping and BRCA-ODX stratification, XMAG exceeded comparison models in F1-score metrics. For RCC subtyping, XMAG achieved an F1-score of 0.916, outperforming Phikon by +0.104 (p<0.01) while showing comparable performance to UNI2 (+0.003, p>0.05). In BRCA-ODX stratification, XMAG demonstrated F1-scores of 0.844, with non-significant improvements over both Phikon (+0.039, p>0.05) and UNI2 (+0.009, p>0.05). These results indicate that XMAG achieved competitive performance across several WSI-level benchmarks while substantially reducing computational overhead. However, performance patterns varied across tasks. XMAG was comparable to 20x foundation models on NSCLC subtyping and RCC subtyping, showed favorable accuracy and F1-score but slightly lower AUC on the BRCA-ODX external cohort, and underperformed 20x models on ISUP grading. These findings suggest that the benefit of 5x cross-magnification distillation is task-dependent. More detailed comparison results can be found in Supplementary Table S3-6.

Compared with the matched 5x DINOv2 baseline, XMAG generally achieved significantly higher performance across benchmark tasks, demonstrating the benefit of transferring high-magnification knowledge through cross-magnification distillation. Specifically, XMAG improved AUC by +0.029 (p<0.001) for NSCLC subtyping, +0.005 (p<0.001) for ISUP grading, and +0.041 (p<0.001) for BRCA-ODX stratification compared with the 5x DINOv2 baseline.

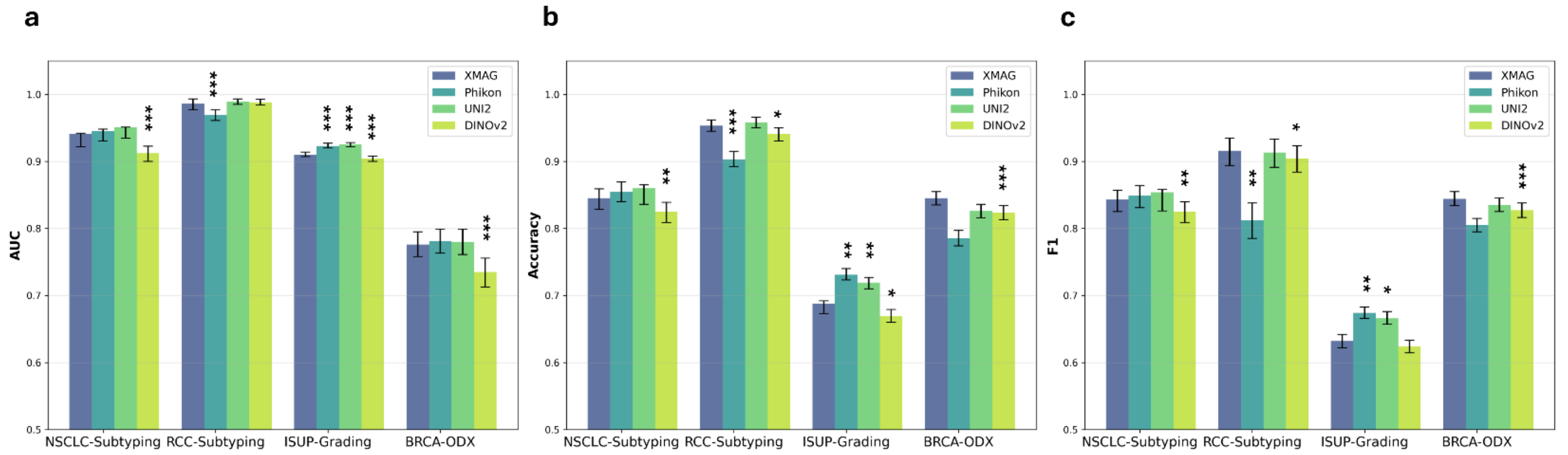

Figure 2. Performance on WSI-level classification tasks. **a-c.** AUC, accuracy, and F1-score comparisons between XMAG, Phikon, UNI2, and DINOv2 across four WSI classification benchmarks. Error bars represent 95% confidence intervals. Statistical significance is indicated by asterisks (*p<0.05, **p<0.01, ***p<0.001).

### 3.2. Speed and scale comparison

The practical deployment of FMs in clinical pathology workflows demands not only high diagnostic accuracy but also computational efficiency compatible with real-time clinical decision-making. We systematically evaluated the computational performance of XMAG against established pathology FMs to assess clinical deployment feasibility.

The reduction in patch number can be described from both theoretical and empirical perspectives. From geometric area scaling, decreasing magnification from 20x to 5x reduces the linear resolution by a factor of four in each spatial dimension; therefore, one 5x patch covers the same tissue area as sixteen 20x patches. This gives a theoretical 16-fold reduction in the number of patches required to cover the same tissue region. In practice, the realized reduction depends on WSI tissue-mask geometry, tissue boundary effects, slide size, and patch extraction parameters. Across the four WSI-level benchmark datasets, 5x processing required an average of 554 patches per WSI, whereas 20x processing required an average of 6,260 patches per WSI, corresponding to an empirical 11.3-fold average reduction (see Figure 3a).

To quantify processing efficiency under a controlled condition, we simulated WSI analysis using representative patch counts derived from our benchmarking datasets: 554 patches of 224×224 pixels for 5× magnification (XMAG) and 6,260 patches for 20× magnification (comparison models). These values represent the averaged patch requirements across the four benchmark datasets at their respective magnifications, reflecting typical coverage differences in our experimental setting. We constructed CPU-generated dummy slides consisting of the corresponding number of image patches for each magnification setting. This benchmark was designed to measure patch-encoding throughput under a controlled tensor-based setting, without including tissue masking, patch I/O, or ABMIL aggregation.

Under identical hardware conditions, XMAG demonstrated remarkable computational efficiency, requiring only 6.82 seconds for patch processing compared to 54.21 seconds for Phikon and 201.25 seconds for UNI2 (Figure 3b). This represents a 7.95-fold speedup over Phikon and a 29.51-fold acceleration compared to UNI2. Translating these processing times to clinical throughput metrics reveals the practical significance of these efficiency gains. XMAG achieves a processing rate of 8.80 WSIs per minute, substantially exceeding both Phikon (1.11 WSIs/minute) and UNI2 (0.30 WSIs/minute).

To contextualize these efficiency gains within the performance-accuracy landscape, we constructed a comprehensive comparison incorporating model parameters, processing speed, and average diagnostic performance across all four WSI-level classification tasks (Figure 1d). This analysis reveals XMAG's unique position in the efficiency-accuracy trade-off space: while maintaining 86M parameters—identical to Phikon—XMAG achieves nearly equivalent diagnostic performance (average AUC across NSCLC, RCC, ISUP, and BRCA-ODX tasks: 0.910 vs 0.913) with dramatically superior throughput. Compared to the larger UNI2 model (682M parameters, average AUC: 0.917), XMAG retains comparable accuracy while delivering 29.51-fold faster processing. Furthermore, because the simulated benchmark isolates patch-encoding throughput and does not include non-model-specific components of a practical WSI pipeline, we further performed an end-to-end runtime benchmark and presented the result in Supplementary Section 1. This analysis showed that XMAG retained substantial acceleration in the encoder forward-pass component. On the other hand, the end-to-end speedup was partially diluted by model-independent overhead from patch I/O, tissue masking, and other pipeline stages.

Together, these results show that cross-magnification distillation can substantially reduce the computational cost of pathology FM inference while maintaining competitive downstream performance. The additional end-to-end benchmark further indicates that practical deployment efficiency depends not only on model inference speed, but also on system-level factors such as patch I/O and pipeline integration.

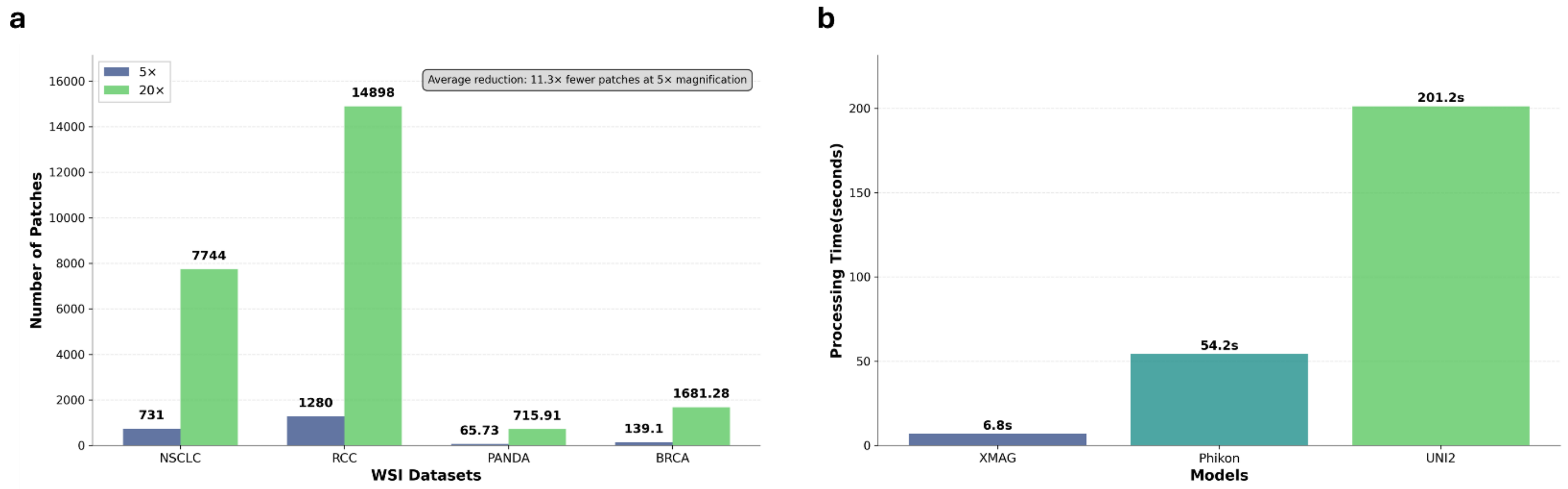

Figure 3. Computational efficiency of 5x and 20x models. **a**. Patch count comparison across WSI datasets showing 11.3-fold average reduction in patches required for 5x versus 20x magnification analysis. **b**. Processing time comparison between XMAG, Phikon, and UNI2 on simulated WSIs.

### 3.3. End-to-end integration enables task-specific adaptation

To evaluate whether joint optimization of FM and MIL components could enhance diagnostic performance, we implemented an end-to-end training framework that integrates XMAG with ABMIL (e2e-XMAG). This approach enables gradient flow from the MIL objective back through the FM[18], allowing task-specific feature adaptation—a capability precluded by conventional frozen encoder pipelines due to memory constraints.

We benchmarked e2e-XMAG on prostate cancer ISUP grading, the most challenging task in our evaluation suite where XMAG exhibited the largest performance gap relative to 20x FMs. The end-to-end framework incorporates gradient checkpointing and selective blocks unfreezing to manage memory requirements while enabling targeted adaptation of high-level feature representations.

e2e-XMAG demonstrated substantial performance improvements over the conventional frozen encoder approach (Table 1). Specifically, end-to-end training achieved increased AUC of 0.924 (+0.014, $p<0.001$), accuracy of 0.713 (+0.025, $p<0.001$), and F1-score of 0.654 (+0.022, $p<0.001$). These gains represent statistically significant improvements across all metrics, indicating that task-specific adaptation effectively compensates for the reduced model capacity relative to larger FMs. Due to the use of gradient checkpointing and partial backbone fine-tuning, the performance gain comes with increased training cost, but without increasing inference-time latency.

To explore the numbers of trainable blocks' impact on the performance, we performed an ablation study to unfreeze different numbers of blocks of the e2e-XMAG backbone (DINOv2-ViT-B) during the end-to-end training (Table S10). We found that unfreezing two backbone blocks yielded the best overall performance. This suggests that selectively unfreezing a small number of high-level transformer blocks is important for balancing task-specific adaptation and training stability. Overall, these results suggest that selective end-to-end optimization further extends XMAG by enabling task-specific adaptation while retaining its computational advantages, providing a direction for boosting efficient foundation-model-based WSI analysis.

Table 1. Performance comparison on prostate cancer ISUP grading (PANDA dataset) between end-to-end XMAG and conventional MIL pipelines with frozen foundation models. Values represent mean ± standard deviation across five-fold cross-validation. P-values in parentheses compare each model with e2e-XMAG.

| | AUC | Accuracy | F1-score |
|---|---|---|---|
| XMAG-ABMIL | 0.910±0.003 ($p<0.001$) | 0.688±0.010 ($p<0.001$) | 0.632±0.012 ($p<0.001$) |
| Phikon-ABMIL | 0.923±0.002 ($p>0.1$) | 0.731±0.009 ($p<0.001$) | 0.674±0.017 ($p<0.001$) |
| UNI2-ABMIL | 0.925±0.004 ($p>0.1$) | 0.719±0.010 ($p>0.1$) | 0.666±0.010 ($p=0.02$) |
| e2e-XMAG | 0.924±0.002 | 0.713±0.005 | 0.654±0.006 |

## 3.4. Patch-level classification and segmentation

Histopathological phenotyping—the identification and classification of distinct tissue morphologies—is fundamental to pathological diagnosis, with pathologists routinely examining tissue architecture across multiple magnifications to characterize cellular and structural phenotypes. In CPath, this critical diagnostic process is typically implemented through patch-level classification, where image patches extracted from WSIs are systematically categorized into distinct phenotypic classes. This task serves as a critical benchmark for evaluating whether FMs can extract semantically meaningful features with genuine histological relevance.

We assessed the tissue phenotyping performance of XMAG using two established patch-level classification datasets: SPIDER-Breast, SPIDER-CRC [29]. These datasets provide comprehensively annotated patches extracted from breast and colorectal cancer WSIs, respectively. Each patch is labeled at 224×224 pixels under 20x magnification, with the surrounding 1120×1120 pixel region available. To simulate 5x magnification patches for XMAG evaluation, we extracted 896×896 pixel regions centered on each patch and downscaled them to 224×224 pixels. SPIDER-Breast contains eighteen distinct tissue categories, while SPIDER-CRC encompasses thirteen categories.

We conducted linear probing experiments to benchmark XMAG against two state-of-the-art 20x FMs: UNI2 and Phikon, and a 5x DINOv2 baseline. For each model, we appended a trainable linear classifier while keeping the FM weights frozen during training, ensuring fair comparison of learned representations.

On the official testing set, XMAG achieved AUC value of 0.988 (95% CI 0.988-0.989) for SPIDER-Breast. In comparison, Phikon achieved an AUC of 0.992 (+0.004, $p>0.1$) and UNI2 achieved an AUC of 0.994 (+0.006, $p=0.05$). On the SPIDER-CRC, XMAG achieved AUC value of 0.990 (95% CI 0.990-0.991). In comparison, Phikon achieved an AUC of 0.994 (+0.004, $p=0.011$) and UNI2 achieved an AUC of 0.994 (+0.004, $p=0.039$). While these results represent a modest decrease compared to the 20x FMs. This modest performance reduction likely stems from label-irrelevant phenotypic information in the peripheral regions of 5x patches. SPIDER dataset annotations are based on the central 224×224 region of each patch, while the surrounding tissue areas incorporated when downscaling from 20x to 5x magnification may contain phenotypes unrelated to the assigned label. More detailed results can be found in Figure 4 and Supplementary Table S7 and Table S8.

In addition to patch-level phenotyping, we further evaluated XMAG's dense prediction performance using colonoscopy lesion segmentation on the DigestPath dataset [30]. Unlike patch-level classification, which relies on overall <CLS> embedding, segmentation requires the encoder to provide spatially resolved local tokens for dense prediction. Following the UNI segmentation evaluation protocol, we used Mask2Former [31] with different pretrained vision encoders and performed WSI-level inference by stitching patch-level predictions. XMAG achieved the highest Dice score among the evaluated encoders, suggesting that the distilled 5x representation retained useful spatial information for this lesion segmentation setting (Figure 4d). Implementation details and additional results can be found in Supplementary Section 2 and Table S9.

To visualize the learned feature representations, we applied t-SNE (t-distributed stochastic neighbor embedding) [32] dimensionality reduction to XMAG and UNI2's patch embeddings (Figure 5). The resulting two-dimensional embedding space revealed distinct clustering of patches according to their phenotypic categories, indicating that XMAG successfully captures histologically relevant features despite operating at reduced magnification. The clear separation between tissue types demonstrates that the cross-magnification distillation framework preserves critical discriminative information necessary for accurate phenotypic classification.

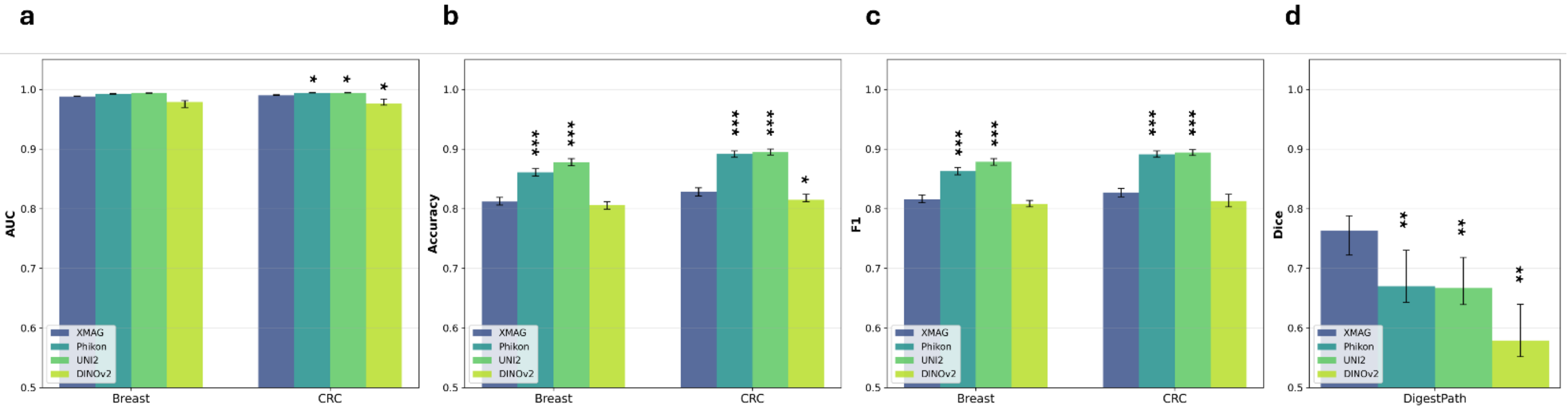

Figure 4. Performance on patch-level classification and segmentation tasks. **a-c.** AUC, accuracy, and F1-score comparisons between XMAG, Phikon, UNI2, and DINOv2 across two patch-level classification benchmarks. **d.** Dice-score comparison on DigestPath dataset. Error bars represent 95% confidence intervals. Statistical significance is indicated by asterisks (*p<0.05, **p<0.01, ***p<0.001).

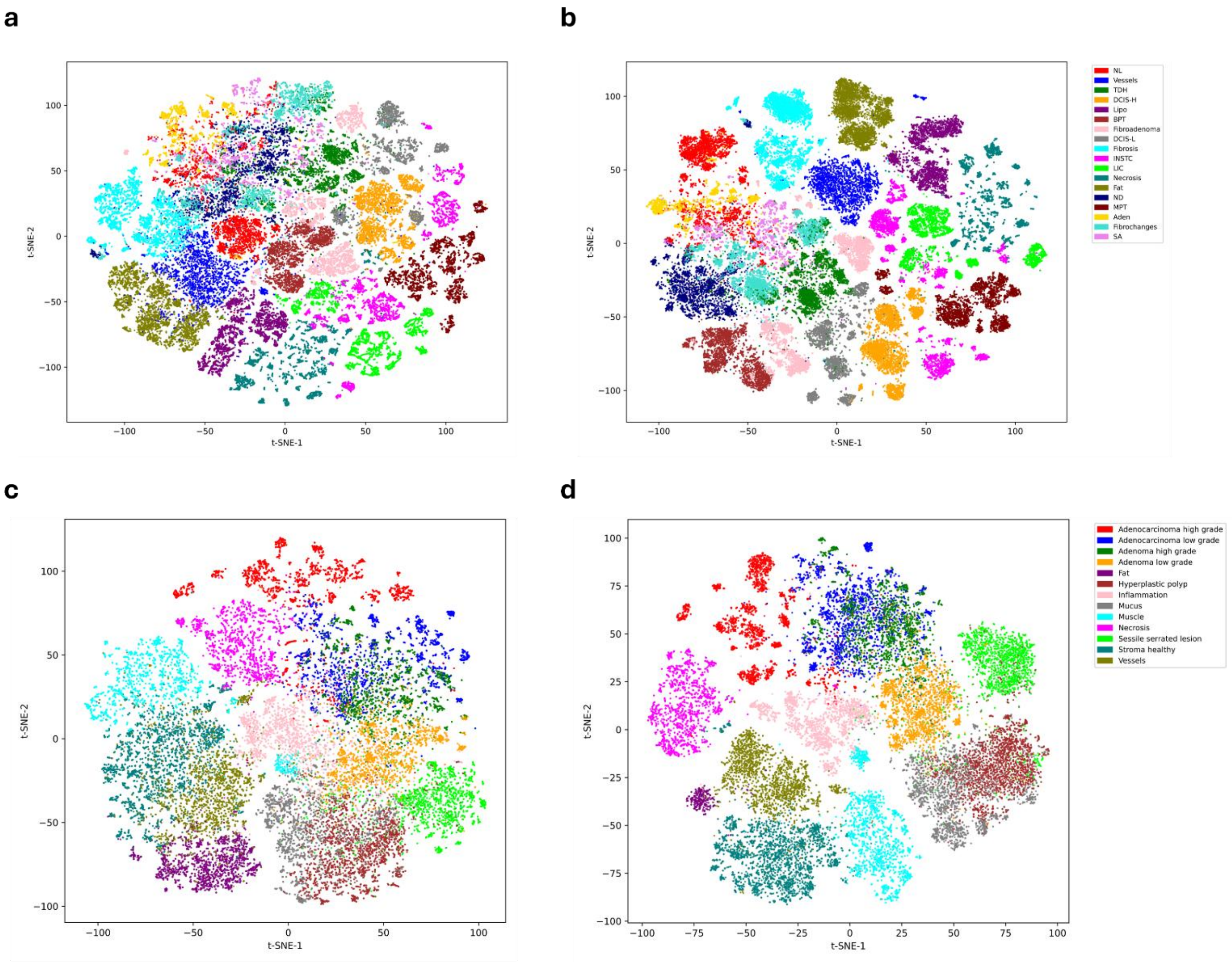

Figure 5. Two-dimensional feature space visualization of XMAG and UNI2 using t-SNE. Patches from the SPIDER dataset are embedded using the XMAG and UNI2 under the inference mode separately, and then mapped to two-dimensional space using the t-SNE technique. **a-b.** Visualization of patch features from SPIDER-Breast dataset encoded by XMAG and UNI2, respectively. **c-d.** Visualization of patch features from SPIDER-CRC dataset encoded by XMAG and UNI2, respectively.

### 3.5. Ablation studies

To isolate the contribution of the proposed cross-magnification distillation framework, we first compared full XMAG with three controlled variants: a vanilla 5× DINOv2 student trained without teacher-guided distillation, a local-only distillation model, and a global-only distillation model. As shown in Table 2, full XMAG consistently outperformed the vanilla 5x DINOv2 baseline on the WSI-level classification tasks, indicating that the observed gains cannot be attributed solely to the 5x ViT-B architecture or the scale of the pretraining dataset. Instead, the results support the added value of teacher-guided cross-magnification knowledge transfer. The loss-component ablation further demonstrated the complementary roles of global and local alignment. Global alignment serves as the primary supervisory signal because it directly supervises the student's <CLS> token, which is used as the tile-level embedding in downstream MIL classification. In contrast, local alignment acts as an auxiliary spatial guidance term that encourages the student to learn region-level information from the 20x teacher. Full XMAG achieved the best overall performance compared with both the global-only and local-only variants, indicating that the primary global alignment objective and the auxiliary local alignment objective both contribute to the final model performance. These results support the combined use of global and local distillation in the proposed framework.

Table 2. Ablation study of cross-magnification distillation design. Vanilla 5x DINOv2, local-only distillation, global-only distillation, and XMAG were evaluated on NSCLC-subtyping and ISUP-grading. Results are reported as mean±standard deviation across five-fold cross-validation. P-values in parentheses compare each ablation with XMAG.

| | NSCLC-subtyping | | ISUP-grading | |
|---|---|---|---|---|
| | AUC | Accuracy | AUC | Accuracy |
| DINOv2 | 0.912±0.016 (p<0.001) | 0.825±0.029 (p<0.01) | 0.904±0.001 (p<0.001) | 0.669±0.008 (p=0.042) |
| $XMAG_{local}$ | 0.770±0.030 (p<0.001) | 0.719±0.047 (p<0.001) | 0.808±0.009 (p<0.001) | 0.477±0.024 (p<0.001) |
| $XMAG_{global}$ | 0.923±0.016 (p<0.01) | 0.839±0.041 (p>0.1) | 0.895±0.005 (p<0.001) | 0.664±0.016 (p<0.001) |
| XMAG | **0.941±0.015** | **0.845±0.033** | **0.910±0.003** | **0.688±0.010** |

To evaluate whether average pooling in the local alignment branch may suppress sparse diagnostic features, we compared three spatial aggregation strategies: average pooling, max pooling, and attention-based pooling. As shown in Table 3, average pooling achieved comparable or superior downstream performance compared with max pooling and attention pooling. This suggests that, for cross-magnification distillation, average pooling provides a stable region-level representation that better matches the semantic granularity of the teacher embedding. In contrast, max pooling may overemphasize isolated high-activation tokens, including artifacts, while attention pooling introduces additional trainable parameters that may not provide consistent benefit under the current distillation setting.

Table 3. Ablation of spatial token aggregation strategies in the local alignment branch. Results are reported as mean±standard deviation across five-fold cross-validation. P-values in parentheses compare each ablation with Mean-pooling.

| | NSCLC-subtyping | | ISUP-grading | |
|---|---|---|---|---|
| | AUC | Accuracy | AUC | Accuracy |
| Max-pooling | 0.901±0.013 (p<0.001) | 0.812±0.039 (p<0.001) | 0.888±0.006 (p<0.001) | 0.649±0.010 (p<0.001) |
| Attention-pooling | 0.919±0.016 (p<0.001) | 0.826±0.035 (p<0.01) | 0.892±0.006 (p<0.001) | 0.654±0.009 (p<0.001) |
| Mean-pooling | **0.941±0.015** | **0.845±0.033** | **0.910±0.003** | **0.688±0.010** |

We further conducted a sensitivity analysis on the weighting coefficients of the global and local distillation losses (Table S11). The default setting, $\lambda_{global} = 1$ and $\lambda_{local} = 0.5$, achieved the best overall performance, supporting the use of global alignment as the primary supervisory signal with moderate local alignment as complementary spatial guidance.

### 3.6. Qualitative analysis

To better understand whether XMAG preserves morphologically meaningful features, we further performed a nearest-neighbor retrieval analysis on the two ROI-level phenotyping datasets using frozen XMAG embeddings. For each query images, the top-5 nearest images were retrieved based on cosine similarity in the embedding space, and the retrieval outcome was summarized by majority voting among the retrieved neighbors. Figure 6 shows representative successful and failed retrieval examples from the SPIDER-Breast dataset. In successful cases, XMAG retrieved images with visually similar histological patterns, suggesting that the learned embedding space groups morphologically related tissue phenotypes together. In failed cases, the retrieved images often still shared partial structural similarity with the query. For example, adenosis images were sometimes retrieved with other glandular structures, such as sclerosing adenosis. Similarly, vessel images could retrieve images containing red blood cells, as shown by the top-1 retrieved image in the bottom row of Figure 6. These examples suggest that some retrieval errors may reflect overlapping histological morphology or mixed tissue composition rather than entirely unrelated feature representations. Additional details of the retrieval analysis and further examples are provided in Supplementary Section 2.

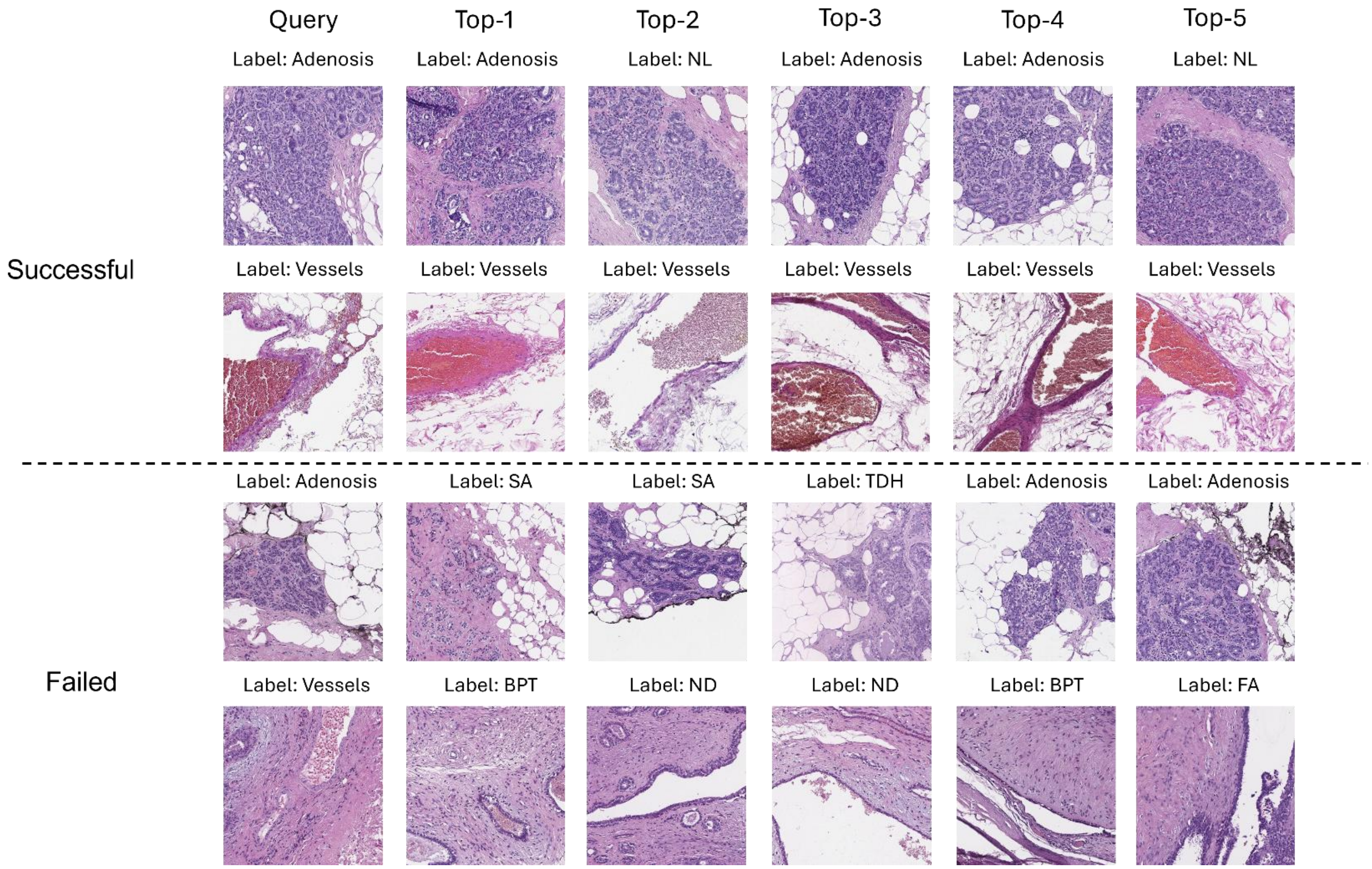

Figure 6. KNN retrieval examples on SPIDER-Breast using XMAG embeddings. For each query image, the top-5 nearest neighbors were retrieved by cosine similarity in the frozen XMAG feature space. Each row shows one query followed by its retrieved neighbors. Abbreviations: NL (Normal Lobules), SA (Sclerosing Adenosis), TDH (Tubular Ductal Hyperplasia), BPT (Benign Phyllodes Tumor), ND (Normal Ducts), FA (Fibroadenoma)

To qualitatively assess whether XMAG preserves diagnostically relevant spatial information, we visualized ABMIL attention maps on representative PANDA slides and compared the attention patterns between XMAG and UNI2. Because prostate cancer grading relies on localized glandular architecture and cytological patterns that determine Gleason scoring and ISUP grade, PANDA provides a suitable setting for evaluating whether the low-magnification XMAG representation can still support localization of relevant tissue regions. As shown in Supplementary Figure S4, although minor differences were observed between XMAG and UNI2 attention maps due to differences in magnification and patch scale, the two models showed broadly consistent attention patterns in the representative examples. Many high-attention regions were spatially consistent between the two models and were frequently localized to glandular tumor areas.

## 4. Discussion

The gap between CPath research and practical deployment remains a major challenge for pathology AI [33]. Recent advances in artificial intelligence have been driven by the emergence of foundation models trained on large-scale datasets, which have demonstrated remarkable performance and transferability across domains such as natural language processing [34], computer vision [35], and multimodal learning [36]. Although pathology foundation models have achieved strong performance across diverse tasks, their high-magnification processing requirements and large parameter counts impose substantial computational costs. These costs are particularly relevant for large-scale WSI analysis and for institutions that require local processing because of data privacy or infrastructure constraints. Our results suggest that cross-magnification distillation provides a practical strategy for reducing the computational burden of pathology foundation models. XMAG achieved substantially faster WSI processing while maintaining competitive performance across several benchmark tasks.

The technical innovation of cross-magnification distillation addresses a counterintuitive challenge in computational pathology: that detailed morphological information captured at high magnification can be effectively transferred to models operating at substantially lower resolution. This finding challenges the prevailing assumption that competitive diagnostic performance necessitates high-magnification analysis with its associated computational overhead. Our dual-level alignment approach—combining global feature matching with local spatial correspondence—enables the transfer of fine-grained morphological knowledge while operating with 16-fold fewer patches per WSI. This approach builds upon the broader knowledge distillation [15] and ViT [37] design literature but introduces novel spatial alignment mechanisms specifically designed for the multi-scale nature of histopathological analysis. The success of this cross-resolution knowledge transfer suggests that much of the diagnostic information captured by high-magnification models can be preserved through appropriately designed distillation frameworks.

A critical advantage of our approach lies in its data efficiency and accessibility. XMAG achieves competitive performance using a training dataset of 3.49 million patches curated entirely from publicly available sources, that is substantially smaller than the proprietary datasets typically required for state-of-the-art models [3]. This data efficiency stems from the knowledge distillation framework, which leverages representations learned by teacher models trained on larger datasets, effectively democratizing access to high-performance pathology AI. This contrasts sharply with the current trend toward increasingly large, private datasets that create barriers to entry for academic institutions and smaller research groups. Our publicly curated dataset and distillation methodology provide a foundation for broader community engagement in CPath research.

The computational efficiency of XMAG enables enhanced training flexibility for downstream applications. While full end-to-end optimization of FM with MIL remains challenging due to memory constraints, XMAG's reduced computational footprint allows for selective fine-tuning approaches using gradient checkpointing techniques. Our experiments on prostate cancer ISUP grading demonstrate that partial end-to-end optimization can yield meaningful performance improvements. This suggests potential for more sophisticated training strategies that could further improve task-specific performance as memory-efficient training techniques continue to advance.

The appropriate use of XMAG should be considered in a task-dependent manner. Based on our results, XMAG is most suitable for efficient preliminary screening, case triage, and high-throughput slide-level prediction tasks where regional tissue architecture or broader morphological patterns provide sufficient diagnostic signal. Examples include tumor subtyping, risk stratification, and lesion segmentation tasks in which larger morphological structures and spatial context are informative. These scenarios may benefit from the substantially reduced patch count and faster inference enabled by 5x analysis, particularly when large numbers of WSIs need to be processed. Tasks that require fine nuclear morphology, or detailed glandular assessment may still benefit from high- or multi-magnification strategies. This is reflected in the performance gap observed on prostate cancer ISUP grading and ROI-level phenotyping, where diagnostic decisions depend more strongly on high-resolution morphological details. Therefore, XMAG is better positioned as an efficient low-magnification model for screening, triage, and broader morphology-driven prediction, while high-magnification models remain important for fine-grained grading and detailed phenotyping.

Several limitations warrant consideration in interpreting these results. First, our evaluation reveals that certain diagnostic tasks may inherently require high-magnification analysis, as evidenced by the performance gap in prostate cancer grading where cellular-level morphological details are critical for accurate ISUP scoring. This suggests that future approaches may need to incorporate multi-resolution strategies that dynamically select appropriate magnification levels based on diagnostic requirements [38]. Second, current memory-efficient training techniques still require trade-offs between computational efficiency and full end-to-end optimization, indicating that further advances in training methodologies will be needed to fully realize the potential of joint foundation model-MIL optimization. Third, our end-to-end runtime analysis showed that practical WSI deployment efficiency is influenced not only by encoder inference speed, but also by pipeline-level factors such as patch I/O, tissue masking, and coordinate extraction. Although XMAG substantially reduces the model-dependent encoding cost, further system-level optimization of data loading and WSI processing will be needed to fully realize the efficiency advantage of XMAG in deployment settings.

Another limitation of our benchmark is that comparisons with existing pathology foundation models are not fully controlled for magnification, patch size, architecture, and pretraining strategy. UNI2 and Phikon were evaluated under their original 20x configurations to provide practical reference benchmarks, whereas XMAG was designed for 5x analysis. Because mature publicly available 5x pathology foundation models are currently limited, we added a matched vanilla 5x DINOv2 baseline using the same ViT-B backbone, input magnification, and pretraining dataset as XMAG. Future studies should further compare XMAG with additional low-magnification pathology foundation models as they become available.

In summary, XMAG introduces a cross-magnification distillation framework for developing efficient low-magnification pathology foundation models. By transferring representations from a 20x teacher to a compact 5x student, XMAG reduces patch requirements and computational cost while maintaining competitive performance across several WSI-level and patch-level benchmarks. Our results also show that the benefits of low-magnification modeling are task-dependent, and applications requiring fine nuclear detail or highly localized morphology may still require high-magnification or multi-resolution approaches. Overall, XMAG provides a scalable and efficient foundation for computationally constrained pathology AI workflows, supporting broader use of foundation-model-based analysis in research and translational settings.

**Datasets Used in the Study**

We evaluated XMAG across multiple publicly available and internal datasets spanning lung, renal, breast, colon, and prostate cancers.

- **CPTAC-NSCLC:** The CPTAC-NSCLC dataset is used for training and testing the NSCLC subtyping model. It provides matched proteogenomic and histopathology data, accessible via the GDC portal.

After excluding normal cases, the dataset includes 668 LUAD slides and 306 LUSC slides. Data are publicly available via the GDC portal (https://portal.gdc.cancer.gov).

- **TCGA-RCC:** The TCGA-RCC dataset is used for training the RCC subtyping model. We employed the TCGA Kidney collections (*KICH, KIRC,* and *KIRP*), comprising 903 diagnostic WSIs from renal cell carcinoma patients. It includes 118 ChRCC slides, 288 PRCC slides, and 497 CCRCC slides. Data are publicly available via the GDC portal (https://portal.gdc.cancer.gov).
- **DHMC-RCC:** The DHMC-RCC dataset was used as an external testing cohort for RCC subtyping. It consists of 563 WSIs from renal cell carcinoma patients, including 23 ChRCC, 101 PRCC, and 344 CCRCC slides. This dataset was held out exclusively for testing and is available at https://bmirds.github.io/KidneyCancer/.
- **TCGA-BRCA:** The TCGA-BRCA dataset is used for training the BRCA-ODX stratification model. consisting of 1065 diagnostic WSIs with corresponding labels. It includes 312 high-risk slides and 753 low-risk slides. It is publicly accessible through the GDC portal (https://portal.gdc.cancer.gov).
- **OSUWMC-BRCA:** OSUWMC-BRCA dataset is an internal dataset that we used for testing the BRCA-ODX stratification model. It includes 147 high-risk slides and 847 low-risk slides. This dataset is not publicly available.
- **PANDA:** The PANDA dataset is used for training and testing the ISUP-grading model. It is one of the largest public resources for prostate cancer grading, includes 10,616 digitized prostate biopsy WSIs collected from multiple international centers. It is publicly available at https://www.kaggle.com/c/prostate-cancer-grade-assessment.
- **SPIDER-CRC:** The SPIDER-CRC is used for training and testing the CRC patch-level phenotyping model. It includes 63,989 training images and 13,193 testing images spanning 13 categories. It is publicly available at https://huggingface.co/datasets/histai/SPIDER-colorectal.
- **SPIDER-Breast:** The SPIDER-Breast is used for training and testing the breast cancer patch-level phenotyping model. It includes 80,858 training images and 12,034 testing images spanning 18 categories. It is publicly available at https://huggingface.co/datasets/histai/SPIDER-breast.
- **DigestPath:** The DigestPath colonoscopy tissue segmentation dataset is used for training and testing the lesion segmentation model. We use the positive cohort, where tumor lesions are present and pixel-level annotations are available, and formulate the task as binary tumor-versus-background segmentation. This cohort contains 250 large WSI-region images from 93 WSIs. It is publicly available at https://digestpath2019.grand-challenge.org/.

**Data Availability**

All TCGA and CPTAC data are publicly available at https://portal.gdc.cancer.gov. DHMC-RCC dataset is available at https://bmirds.github.io/KidneyCancer/. PANDA dataset is available at https://www.kaggle.com/c/prostate-cancer-grade-assessment. SPIDER datasets are available at https://huggingface.co/histai/datasets. DigestPath is available at https://digestpath2019.grand-challenge.org/.

**Code Availability**

Our code and model weights are made available on https://github.com/AI4Path-Lab/XMAG.

**Funding**

The project described was supported in part by R01 CA276301 (PIs: Niazi and Chen) from the National Cancer Institute. The project was also supported partly by Pelotonia under IRP CC13702 (PIs: Niazi, Vilgelm, and Roy), The Ohio State University Department of Pathology and Comprehensive Cancer Center. The content is solely the responsibility of the authors and does not necessarily represent the official views of the National Cancer Institute or National Institutes of Health, or Pelotonia, or The Ohio State University.

**Acknowledgements**

The authors thank the members of the AI4PATH Laboratory and the Department of Pathology at The Ohio State University for their valuable discussions and technical assistance. We also acknowledge the support and resources provided by National Institutes of Health, Pelotonia, and The Ohio State University.

**Supplementary materials**

**1. End-to-end WSI runtime benchmark**

The simulated WSI analysis experiment reported in the main results was designed to estimate the theoretical speed advantage of XMAG over high-magnification foundation models during pure patch encoding, where input patches were already represented in tensor format. To provide a more realistic assessment of deployment efficiency, we additionally performed an end-to-end WSI runtime benchmark on a sampled set of 200 OSUWMC-BRCA slides. The evaluated pipeline included tissue masking, coordinate extraction, patch extraction and data loading (I/O), patch encoding, and ABMIL aggregation (see Figure S1). Within the pipeline, tissue masking was the only shared step across all models; all subsequent steps were performed separately for each model. This benchmark therefore captures not only model-dependent inference cost, but also non-model-specific overhead introduced by WSI processing and data loading. Patches are extracted according to models' native inference settings. Runtime was reported component-wise in minutes. All experiments were performed on a single NVIDIA A100 40GB GPU.

As shown in Table S1, XMAG required 39.67 minutes to process 200 slides from tissue masking to slide-level prediction, compared with 207.96 minutes for Phikon and 240.38 minutes for UNI2. This corresponds to a 5.24-fold reduction in end-to-end runtime compared with Phikon and a 6.06-fold reduction compared with UNI2. The largest difference was observed in the encoder forward-pass time. XMAG required 5.00 minutes for encoding, whereas Phikon and UNI2 required 150.70 and 186.84 minutes, respectively, corresponding to 30.14-fold and 37.37-fold longer encoding time than XMAG. These results indicate that the reduced magnification and compact encoder substantially decrease the model-dependent inference cost.

The end-to-end speedup was smaller than the pure encoding speedup observed in the simulated experiment because non-encoder components, particularly patch I/O and coordinate extraction, contributed a non-negligible fraction of the total runtime. Although XMAG reduced the number of extracted patches by operating at 5x magnification, patch I/O still accounted for a large portion of its full pipeline runtime. This suggests that the deployment advantage of low-magnification compact models can be partially diluted by fixed or partially model-independent overheads. Therefore, fully realizing the efficiency benefit of XMAG will require system-level optimization of the WSI processing pipeline, such as efficient patch caching, parallel I/O, optimized tile readers, and tighter integration between coordinate extraction, patch loading, and model inference.

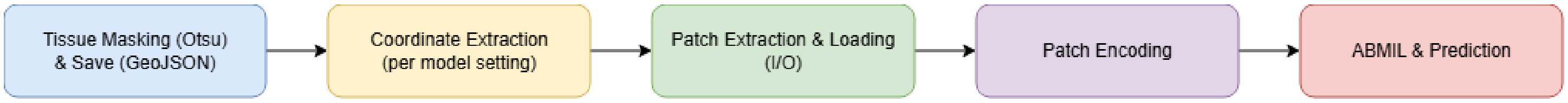

Figure S1. Flowchart of end-to-end WSI processing pipeline.

Table S1. Component-wise end-to-end runtime benchmark on 200 OSUWMC-BRCA slides. All runtime values are reported in minutes. “Patch I/O” includes patch extraction and data loading. “Encoding” denotes the pure encoder forward-pass time. Values in parentheses indicate the fold increase in runtime relative to XMAG.

| **Models** | **Masking** | **Coord. extraction** | **Patch I/O** | **Encoding** | **ABMIL** | **Total** |
|---|---|---|---|---|---|---|
| $\text{UNI2}_{20x, 256}$ | 2.49 | 13.88 (↑ 5.42x) | 37.12 (↑ 1.25x) | 186.84 (↑ 37.37x) | 0.05 (↑ 8.33x) | 240.38 (↑ 6.06x) |
| $\text{Phikon}_{20x, 256}$ | 2.49 | 13.96 (↑ 5.45x) | 40.78 (↑ 1.38x) | 150.70 (↑ 30.14x) | 0.03 (↑ 5x) | 207.96 (↑ 5.24x) |
| $\text{XMAG}_{5x, 224}$ | 2.49 | 2.56 | 29.61 | 5.00 | 0.006 | 39.67 |

## 2. Implementation details of segmentation experiment

We evaluated lesion segmentation on the DigestPath dataset [30] to assess the spatially resolved representations learned by different pretrained vision encoders. We performed five-fold cross-validation and reported the average performance across five folds. For each fold, we constructed the training set by randomly cropping annotated image patches from the training WSIs and trained the segmentation models at the patch level. During testing, predictions were generated on image patches in a sliding-window manner and then stitched back to the WSI coordinate space to obtain slide-level segmentation masks for evaluation. Each model was used in its default input magnification and patch size. For sliding-window inference, we used $224 \times 224$ window size with a stride of 168-pixel, and predictions in overlapping regions were averaged at the probability level.

Following UNI’s segmentation protocol [3], we adopted a Mask2Former-based segmentation pipeline for this experiment. Because ViT backbones produce non-hierarchical feature maps, we incorporated ViT-Adapter [39] to convert ViT features into multi-scale representations before the Mask2Former decoder. The segmentation models were initialized with different pretrained encoders and trained under the same protocol for fair comparison. We used AdamW optimization with an initial learning rate of 1e-4 and a weight decay of 0.05. The learning rate was decayed by a factor of 10 at 90% and 95% of the total training schedule. All models were trained for 50 epochs with a batch size of 16. Performance was evaluated on the validation set every 5 epochs, and the checkpoint with the best validation performance was selected for final testing. Representative qualitative segmentation results are shown in Figure S2, where XMAG predictions are compared with the corresponding ground-truth lesion annotations on DigestPath examples. Quantitative results can be found in Table S9.

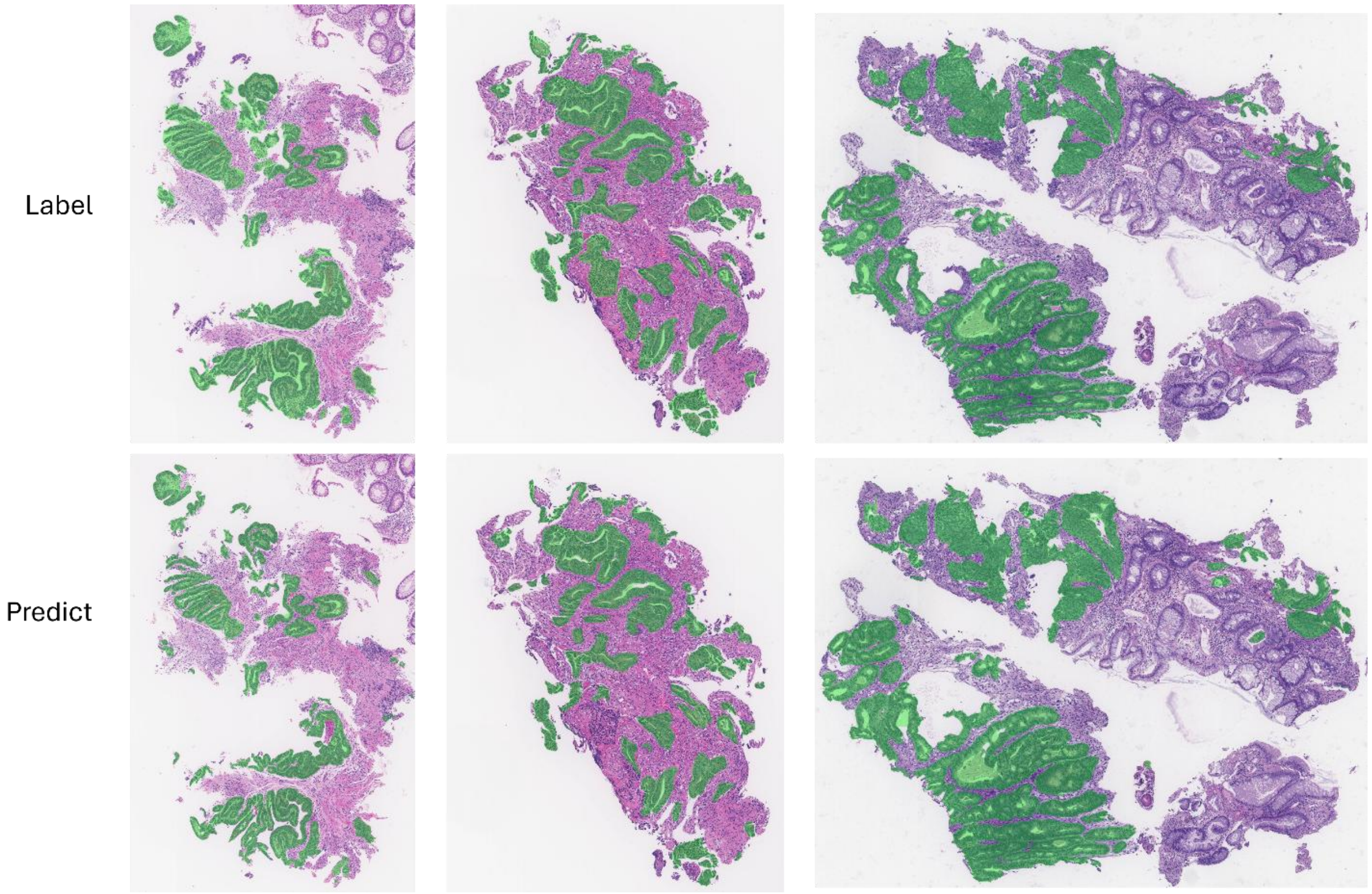

Figure S2. Representative lesion segmentation results of XMAG on the DigestPath dataset. Green regions indicate lesion areas. For each example, the ground-truth annotation and XMAG prediction are shown in separate rows.

## 3. Implementation details of retrieval analysis

We performed nearest-neighbor retrieval analysis on SPIDER-Breast and SPIDER-CRC using the frozen XMAG encoder. No retrieval-specific model was trained. For each dataset, we constructed a class-balanced gallery set from the training split and a class-balanced query set from the test split. Specifically, we randomly sampled 100 images per class as gallery references and 30 images per class as query images. For each image, we extracted the XMAG patch-level embedding and applied L2 normalization. Retrieval was performed using cosine similarity, computed as the dot product between normalized query and gallery embeddings. For each query image, the top-5 nearest gallery images were selected. The retrieval class was assigned by majority voting among the top-5 retrieved references; ties were resolved using the class of the highest-ranked retrieved image among the tied classes.

Figure S3 shows representative successful and failed retrieval examples from the SPIDER-CRC dataset. Overall, the successful cases demonstrate that XMAG retrieves images with similar histological phenotypes in the embedding space. In the failed cases, the retrieved images often still share related morphology with the query. For example, adenoma and adenocarcinoma images may be retrieved together because both are glandular epithelial neoplasms. Similarly, sessile serrated lesions and hyperplastic polyps may be confused due to their overlapping sawtooth-like architecture.

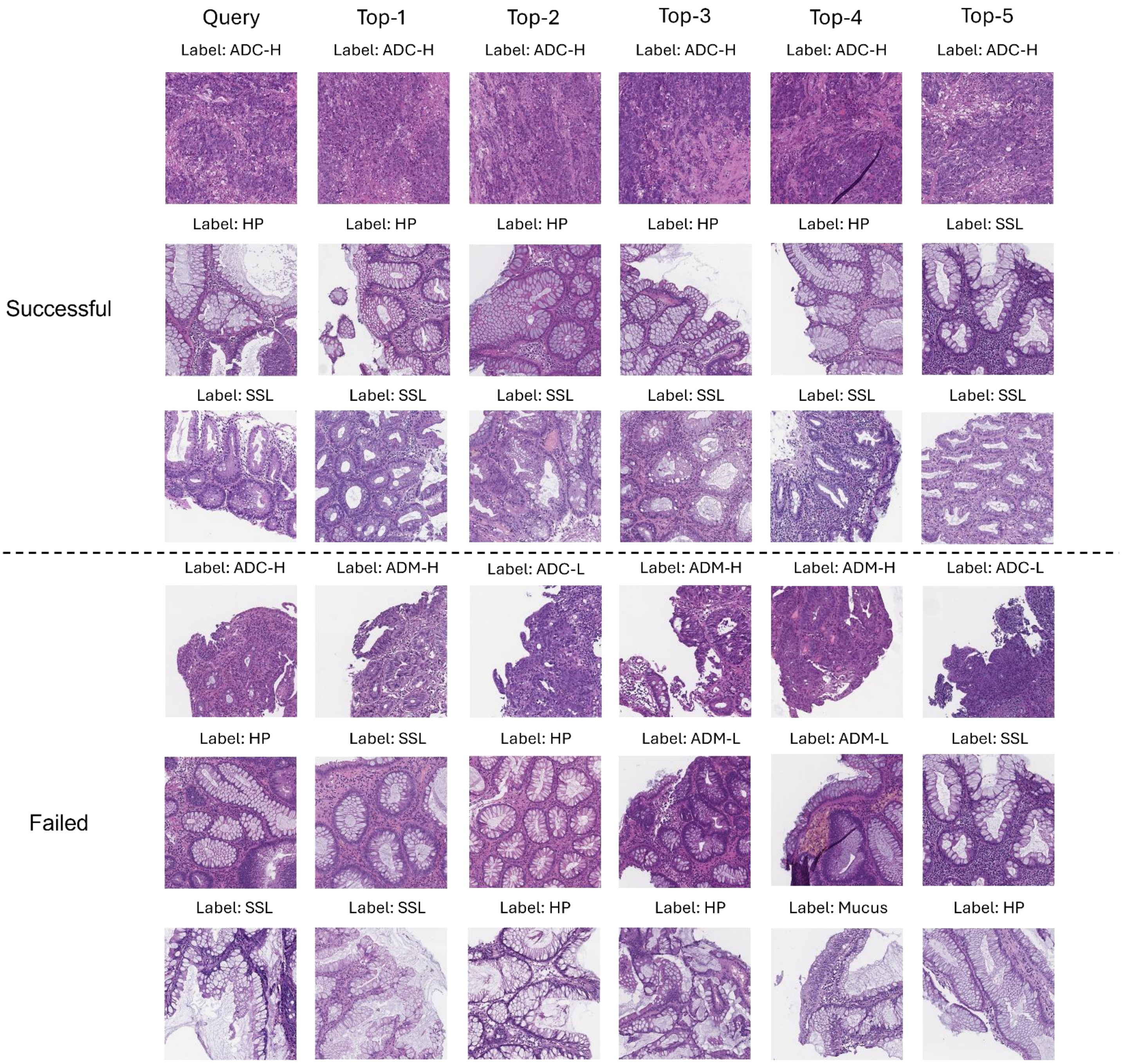

Figure S3. KNN retrieval examples on SPIDER-CRC using XMAG embeddings. For each query image, the top-5 nearest neighbors were retrieved by cosine similarity in the frozen XMAG feature space. Each row shows one query followed by its retrieved neighbors. Abbreviations: ADC-H/L (Adenocarcinoma high/low grade), ADM-H/L (Adenoma high/low grade), HP (Hyperplastic polyp), SSL (Sessile serrated lesion)

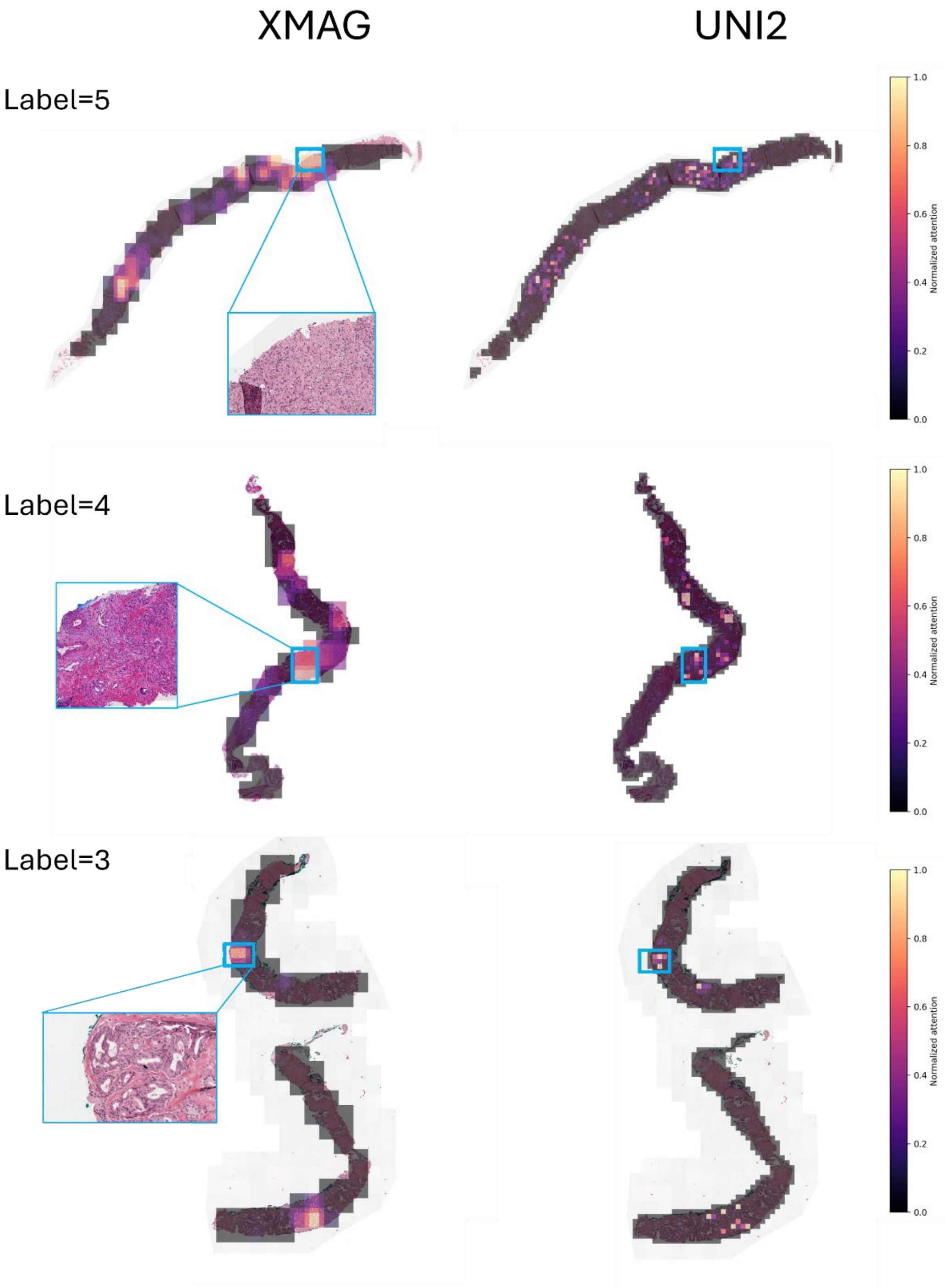

Figure S4. ABMIL attention map visualization on representative PANDA slides. Representative cases from ISUP grades 5, 4, and 3 are shown. For each case, attention maps generated using XMAG and UNI2 features are presented, with zoomed-in views of high-attention regions. The highlighted regions are primarily localized to glandular tumor areas, which are closely related to Gleason scoring and ISUP grade assignment.

Table S2. Summary of statistical tests used for model performance comparisons across classification and segmentation tasks.

| | | Accuracy | AUC/macro-AUC | F1/macro-F1 | Dice |
|---|---|---|---|---|---|
| Binary classification | NSCLC-subtyping | McNemar's test | DeLong's test | Paired bootstrap | - |
| | BRCA-ODX | McNemar's test | DeLong's test | Paired bootstrap | - |
| Multiclass classification | RCC-subtyping | Paired bootstrap | Paired bootstrap | Paired bootstrap | - |
| | ISUP-grading | Paired bootstrap | Paired bootstrap | Paired bootstrap | - |
| | SPIDER-breast | Paired bootstrap | Paired bootstrap | Paired bootstrap | - |
| | SPIDER-crc | Paired bootstrap | Paired bootstrap | Paired bootstrap | - |
| Segmentation | DigestPath | - | - | - | Paired bootstrap |

Table S3. NSCLC subtyping performance of different foundation models on CPTAC-NSCLC dataset. Averaged performance values and standard deviations across five-fold cross-validation are reported. P-values in parentheses compare each model with XMAG.

| | Mag | Model size | Accuracy | F1 | AUC |
|---|---|---|---|---|---|
| Phikon | 20x | 86M | 0.855±0.018 (p>0.1) | 0.849±0.011 (p>0.1) | 0.945±0.011 (p>0.1) |
| UNI2 | 20x | 682M | 0.860±0.013 (p>0.1) | 0.854±0.013 (p>0.1) | 0.951±0.008 (p=0.083) |
| DINOv2 | 5x | 86M | 0.825±0.029 (p<0.01) | 0.825±0.030 (p<0.01) | 0.912±0.016 (p<0.001) |
| XMAG | 5x | 86M | 0.845±0.033 | 0.843±0.023 | 0.941±0.015 |

Table S4. RCC subtyping performance of different foundation models on DHMC dataset. ABMIL models are trained on the TCGA-RCC dataset and tested on the DHMC dataset. Averaged performance values and standard deviations across five-fold cross-validation are reported. P-values in parentheses compare each model with XMAG.

| | Mag | Model size | Accuracy | F1 | AUC |
|---|---|---|---|---|---|
| Phikon | 20x | 86M | 0.903±0.013 (p<0.001) | 0.812±0.017 (p<0.01) | 0.969±0.006 (p<0.001) |
| UNI2 | 20x | 682M | 0.958±0.015 (p>0.1) | 0.913±0.021 (p>0.1) | 0.989±0.003 (p>0.1) |
| DINOv2 | 5x | 86M | 0.941±0.014 (p=0.01) | 0.904±0.011 (p=0.04) | 0.982±0.001 (p>0.1) |
| XMAG | 5x | 86M | 0.953±0.008 | 0.916±0.015 | 0.986±0.009 |

Table S5. BRCA-ODX stratification performance of different foundation models on OSUWMC dataset. ABMIL models are trained on the TCGA-BRCA dataset and tested on the OSUWMC dataset. Averaged performance values and standard deviations across five-fold cross-validation are reported. P-values in parentheses compare each model with XMAG.

| | Mag | Model size | Accuracy | F1 | AUC |
|---|---|---|---|---|---|
| Phikon | 20x | 86M | 0.785±0.068 (p>0.1) | 0.805±0.048 (p>0.1) | 0.781±0.003 (p>0.1) |
| UNI2 | 20x | 682M | 0.826±0.030 (p=0.095) | 0.835±0.022 (p>0.1) | 0.780±0.015 (p>0.1) |
| DINOv2 | 5x | 86M | 0.824±0.028 (p<0.001) | 0.827±0.015 (p<0.001) | 0.735±0.005 (p<0.001) |
| XMAG | 5x | 86M | 0.845±0.016 | 0.844±0.007 | 0.776±0.006 |

Table S6. Prostate cancer ISUP grading performance of different foundation models on PANDA dataset. Averaged performance values and standard deviations across five-fold cross-validation are reported. P-values in parentheses compare each model with XMAG.

| | Mag | Model size | Accuracy | F1 | AUC |
|---|---|---|---|---|---|
| Phikon | 20x | 86M | 0.731±0.009 (p<0.01) | 0.674±0.017 (p<0.01) | 0.923±0.002 (p<0.001) |
| UNI2 | 20x | 682M | 0.719±0.010 (p<0.01) | 0.666±0.010 (p=0.012) | 0.925±0.004 (p<0.001) |
| DINOv2 | 5x | 86M | 0.669±0.008 (p=0.042) | 0.624±0.009 (p>0.1) | 0.904±0.001 (p<0.001) |
| XMAG | 5x | 86M | 0.688±0.010 | 0.632±0.012 | 0.910±0.003 |

Table S7. Breast cancer ROI phenotyping performance of different foundation models on SPIDER-Breast dataset. We performed linear probing based on different models and report the performance based on the official testing set of SPIDER-Breast. P-values in parentheses compare each model with XMAG.

| | Mag | Model size | Accuracy | F1 | AUC |
|---|---|---|---|---|---|
| Phikon | 20x | 86M | 0.861 (p<0.001) | 0.863 (p<0.001) | 0.992 (p>0.1) |
| UNI2 | 20x | 682M | 0.878 (p<0.001) | 0.863 (p<0.001) | 0.994 (p=0.053) |
| DINOv2 | 5x | 86M | 0.806 (p=0.061) | 0.808 (p>0.1) | 0.979 (p=0.074) |
| XMAG | 5x | 86M | 0.812 | 0.816 | 0.988 |

Table S8. CRC ROI phenotyping performance of different foundation models on SPIDER-CRC dataset. We performed linear probing based on different models and reported the performance based on the official testing set of SPIDER-CRC. P-values in parentheses compare each model with XMAG.

| | Mag | Model size | Accuracy | F1 | AUC |
|---|---|---|---|---|---|
| Phikon | 20x | 86M | 0.892 (p<0.001) | 0.891 (p<0.001) | 0.994 (p=0.011) |
| UNI2 | 20x | 682M | 0.895 (p<0.001) | 0.894 (p<0.001) | 0.994 (p=0.039) |
| DINOv2 | 5x | 86M | 0.815 (p=0.032) | 0.813 (p=0.058) | 0.976 (p=0.021) |
| XMAG | 5x | 86M | 0.828 | 0.827 | 0.990 |

Table S9. Colonoscopy lesion segmentation performance of different foundation models on DigestPath dataset. We connected foundation models with Mask2Former architecture for segmentation. Averaged performance values and standard deviations across five-fold cross-validation are reported. P-values in parentheses compare each model with XMAG.

| | Mag | Dice |
|---|---|---|
| Phikon | 20x | 0.670±0.027 (p<0.01) |
| UNI2 | 20x | 0.667±0.030 (p<0.01) |
| DINOv2 | 5x | 0.579±0.051 (p<0.01) |
| XMAG | 5x | 0.764±0.008 |

Table S10. Ablation study of tunable blocks of e2e-XMAG. We trained and tested the performance of e2e-XMAG with varied trainable blocks of the DINOv2-ViT-B backbone on the PANDA dataset.

| Trainable blocks | Accuracy | F1 | AUC |
|---|---|---|---|
| 0 | 0.688±0.010 | 0.632±0.012 | 0.910±0.003 |
| 1 | 0.685±0.005 | 0.611±0.012 | 0.923±0.025 |
| 2 | **0.713±0.005** | **0.654±0.006** | **0.924±0.002** |
| 4 | 0.693±0.009 | 0.628±0.013 | 0.908±0.002 |
| 6 | 0.690±0.007 | 0.629±0.009 | 0.911±0.005 |
| 12 (all blocks) | 0.447±0.008 | 0.346±0.025 | 0.764±0.011 |

Table S11. Ablation of global and local distillation loss weights. Results are reported as mean ± standard deviation across five-fold cross-validation. P-values in parentheses compare each ablation setting with XMAG using the default configuration, $\lambda_g$= 1 and $\lambda_l$= 0.5, where $\lambda_g$ and $\lambda_l$ indicate weights of global and local alignment loss separately.

| | | $\lambda_g = 1,\ \lambda_l = 0.5$ | $\lambda_g = 0,\ \lambda_l = 1$ | $\lambda_g = 1,\ \lambda_l = 0$ | $\lambda_g = 1,\ \lambda_l = 1$ |
|---|---|---|---|---|---|
| NSCLC-subtyping | AUC | **0.941±0.015** | 0.770±0.030 (p<0.001) | 0.923±0.016 (p<0.01) | 0.882±0.033 (p<0.001) |
| | Accuracy | **0.845±0.033** | 0.719±0.047 (p<0.001) | 0.839±0.041 (p>0.1) | 0.785±0.042 (p<0.001) |
| ISUP-grading | AUC | **0.910±0.003** | 0.808±0.009 (p<0.001) | 0.895±0.005 (p<0.001) | 0.872±0.003 (p<0.001) |
| | Accuracy | **0.688±0.010** | 0.477±0.024 (p<0.001) | 0.664±0.016 (p<0.001) | 0.610±0.007 (p<0.001) |